\documentclass{article}
\usepackage[utf8]{inputenc}
\usepackage{spconf,amsmath,graphicx}
\graphicspath{{figures/}}

\usepackage{tikz}
\usetikzlibrary{shapes.geometric, arrows, calc, fit}

\title{Semi-Supervised Learning with Sparse Autoencoders\\ in Phone Classification}
\name{Akash Kumar Dhaka and Giampiero Salvi}
\address{KTH Royal Institute of Technology,\\
School of Computer Science and Communication,\\
Dept. for Speech, Music and Hearing,
Stockholm, Sweden\\
\texttt{\{akashd, giampi\}@kth.se}}
\begin{document}
%
\maketitle
\begin{abstract}
We propose the application of a semi-supervised learning method to improve the performance of acoustic modelling for automatic speech recognition based on deep neural networks. As opposed to unsupervised initialisation followed by supervised fine tuning, our method takes advantage of both unlabelled and labelled data simultaneously through mini-batch stochastic gradient descent.
We tested the method with varying proportions of labelled vs unlabelled observations in frame-based phoneme classification on the TIMIT database. Our experiments show that the method outperforms standard supervised training for an equal amount of labelled data and provides competitive error rates compared to state-of-the-art graph-based semi-supervised learning techniques.

\end{abstract}
\section{Introduction}
\label{sec:intro}
Deep Learning has revolutionised research in Automatic Speech Recognition (ASR) as well as many other fields of application of machine learning (see \cite{LeCunEtAl2015Nature, Schmidhuber2015NeuralNetworks} for extensive reviews). Despite, the recent significant improvements made in word error rates (WERs), most of the experiments have been reported on large fully-labelled data sets. The initial paradigm, where unsupervised initialisation of the network weights was followed by supervised fine-tuning of the parameters \cite{ErhanEtAl2010WhyPreTraining, SutskeverEtAl2013InitializationAndMomentum}, was abandoned in favour of fully supervised methods with more efficient models (e.g. \cite{ZeilerEtAl2013ReLU}). However, for under-resources languages, where large amounts of labelled data are not available, non fully supervised learning techniques are still relevant. Unsupervised learning has the limit of finding an initial set of weights, and consequently data representations, that are not specifically optimised for the problem at hand. As an example, we would find the same representations for speech or speaker recognition which have orthogonal objectives. An alternative learning paradigm, that has recently been applied in the field of computer vision as well as ASR, is semi-supervised learning where labelled and unlabelled observations are used jointly \cite{coates-2011, liu-2013, liu-2014, LiuAndKirchoff2016}.  
Semi-supervised learning using neural network has also been explored in \cite{vesley-2013}, by means of a self-training scheme. The self-training scheme is, however, based on heuristics and prone to reinforcing poor predictions. 


The work done by \cite{liu-2013, liu-2014} is one of the first attempts on using these semi-supervised learning in ASR. The authors propose a number of algorithms employing graph based learning (GBL-SSL), and obtain better WERs over a baseline neural network. In \cite{LiuAndKirchoff2016} the authors extend the initial results from frame based phoneme classification to large vocabulary ASR. Graph based learning is, however, computationally intensive, and the addition of a new point in data requires the reevaluation of the graph laplacian.

In \cite{ranzato-2008}, Ranzato and Szummer propose a semi-supervised learning method based on linearly combining the supervised cost function of a deep classifier with the unsupervised cost function of a deep autoencoder and minimising the combination of costs through mini-batch stochastic gradient descent via standard backpropagation. The authors apply their method to finding representations of text documents for information retrieval and classification.

We propose to use a similar approach to frame-based phoneme recognition in ASR. Although our objective function is the same as the one proposed in \cite{ranzato-2008}, our set up is different in a number of ways. Firstly, instead of the compact and lower dimensional encoding used in \cite{ranzato-2008}, we employ sparse encoding. Secondly, instead of stacking a number of encoders, decoders and classifiers in a deep architecture as in \cite{ranzato-2008}, we use a single layer model. This is motivated by work in \cite{coates-2011}, where the authors analyse the effect of several model parameters in unsupervised learning of neural networks on computer vision benchmark data sets such as CIFAR-10 and NORB. They conclude that state-of-the-art results can be achieved with single layer networks regardless of the learning method, if an optimal model setup is chosen.

We perform phoneme recognition on the TIMIT data set and compare the performance of our model with the results obtained with standard supervised learning and with the computationally more expensive GBL methods.

The paper is organised as follows: Section~\ref{sec:method} describes the method. Section~\ref{sec:experiments} reports details on the experimental set-up. Section~\ref{sec:results} reports the results and, finally, Section~\ref{sec:conclusions} concludes the paper.


\section{METHOD}
\label{sec:method}
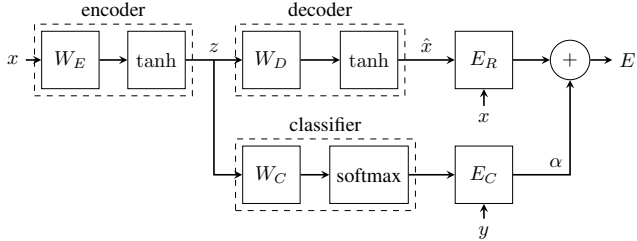
\begin{figure}
\tikzstyle{box} = [rectangle, minimum width=1cm, minimum height=1cm,text centered, draw=black]
\tikzstyle{ball} = [circle, minimum width=5mm, minimum height=5mm,text centered, draw=black]
\tikzstyle{arrow} = [thick,->,>=stealth]
\resizebox{\columnwidth}{!}{%
\begin{tikzpicture}[node distance=2cm]
\node (x) {$x$};
\node (we) [box, right of=x, node distance=1cm] {$W_E$};
\node (ae) [box, right of=we, node distance=1.5cm] {$\tanh$};
\node (wd) [box, right of=ae] {$W_D$};
\node (ad) [box, right of=wd, node distance=1.7cm] {$\tanh$};
\node (er) [box, right of=ad] {$E_R$};
\node (sum) [ball, right of=er, node distance=1.5cm] {$+$};
\node (wc) [box, below of=wd] {$W_C$};
\node (ac) [box, below of=ad] {softmax};
\node (ec) [box, below of=er] {$E_C$};
\node (e) [right of=sum, node distance=1cm] {$E$};
\node (y) [below of=ec, node distance=1cm] {$y$};
\node (xx) [below of=er, node distance=1cm] {$x$};

\draw [arrow] (x) -- (we);
\draw [arrow] (we) -- (ae);
\draw [arrow] (ae) -- node[anchor=south] {$z$} (wd);
\draw [arrow] (wd) -- (ad);
\draw [arrow] (ad) -- node[anchor=south] {$\hat{x}$} (er);
\draw [arrow] (er) -- (sum);
\draw [arrow] (wc) -- (ac);
\draw [arrow] (ac) -- (ec);
\draw [arrow] (ec) -| node[anchor=south east] {$\alpha$} (sum);
\draw [arrow] (sum) -- (e);
\draw [arrow] ($(ae)!0.5!(wd)$) |- (wc);
\draw [arrow] (y) -- (ec);
\draw [arrow] (xx) -- (er);

\node[draw,dashed,fit=(we) (ae), label=encoder] {};
\node[draw,dashed,fit=(wd) (ad), label=decoder] {};
\node[draw,dashed,fit=(wc) (ac), label=classifier] {};
\end{tikzpicture}
} 
\caption{Flow chart for the cost calculation in a single layer of the network. Three components are considered: encoder, decoder, and classifier. The loss is weighted sum of cross-entropy $E_C$ and reconstruction loss $E_R$. If several layers are stacked together, only the encoder/decoder pairs are retained after training.}
\label{fig:model}
\end{figure}

The architecture of a single layer of our model is depicted in Figure~\ref{fig:model}. If we remove the bottom path, this is equivalent to an autoencoder with a set of encoding weights, a logistic layer, a subsequent set of decoding weights and a new non-linearity. In our model, the representation $z$ obtained by the encoder is also used by a classifier in parallel with the regular decoder. The aim of combining unsupervised and supervised cost functions is to use both the unlabelled and labelled data in an efficient way in order to obtain good representations of the input as well as good prediction and discriminative abilities from our network.
 
Although the figure depicts a single layer, in \cite{ranzato-2008} it was shown that a stack of such elements can be trained layer-by-layer in a greedy way. In our experiments, only single layer models were considered.

The model is trained optimising the combined cost of the reconstruction error $E_{R}$ and the classification errors $E_{C}$ given respectively by the autoencoder and the classification network. The combination is linear and defined as:
\begin{equation}
\label{eqn:cost-combined}
E = E_R + \alpha E_C
\end{equation}
where $\alpha$ is a hyper-parameter controlling the proportion of the two costs in the objective function. $\alpha$ is optimised on a validation set that is independent from the training set. Its optimal value depends in general to the proportion of labelled versus unlabelled examples in the training set, as will also be shown in Section~\ref{sec:results}.

In the supervised setting, the cost function is the cross-entropy logloss given by:
\begin{eqnarray}
\label{eqn:cost-classify}
E_C &=& -\sum_{i=1}^{N_C}{y_{i}\log h_{i}}  \\
h_{j} &=& \frac{\exp((W_C)_{j} . z + b_{Cj})}{\sum_{i} \exp((W_{C})_{i}.z + b_{Ci})},
\end{eqnarray}
where $h$ denotes softmax output of the classifier, $W_C$ and $b_C$ are the set of weights and biases for the classification network and $N_C$ is the number of output classes. The variable $z$ denotes the output of the encoder network and is defined as:
\begin{equation}
z = \tanh(W_{E}x + b_{E}),
\end{equation}
Where $W_{E}$ and $b_{E}$ are the weights and biases of the encoder network, and $x$ is the input to the entire model.

In the unsupervised path through the model, the cost function is the second degree norm of difference between original input and reconstructed input, where the input and output have the same dimensions. It has been found, that adding noise to the original input by a process called 'corruption' in which some dimensions of the input vector are randomly picked and set to zero, helps the network to learn even a better representation as described in \cite{Bengio-nips-2006}. 
In this case, the encoded vector $z$ defined above is fed to the decoder layer to produce the final output in the form.
\begin{equation}
\hat{x} = \tanh(W_{D}z + b_{D}) ,
\label{eqn:decoder}
\end{equation}
where 
$W_{D}$ and $b_D$ are the weights and biases of the decoder network. Given the above definitions, the unsupervised reconstruction error $E_R$ is defined as:
\begin{eqnarray}
\label{eqn:cost-reconstruct}
E_R = \sum^{p}_ {i=1} ||x_{i} - \hat{x}_{i} ||^2,
\end{eqnarray}
where $x_{i}$ is the input for a single datapoint, and $\hat{x}_{i}$ is the reconstructed output for a single datapoint. This cost function is the same as that of a regular auto-encoder. In practice, we compute the cost $E_{R}$ averaged over a batch of $p$ points, which is why the optimisation is called as mini-batch Stochastic Gradient Descent (SGD).


When the input datapoint is not accompanied by a label, the classifier part of the layer is not updated, and the loss function simply reduces to $E_{R}$. This model can be iteratively applied to several layers. However, in our experiments, we use just a single layer for feature representation. It is important to note that the update of encoder weights $W_{E}$ is dependent both on the decoder weights $W_{D}$ and on the classifier weights $W_{C}$, and the delta propagated in the backpropagation algorithm will be a linear combination of the deltas calculated in both parts. We used Adaptive Learning Rate scheme with linear decay, in which the learning rate decays linearly after a certain number of epochs.

The size of the hidden representation $z$ is larger than the input size in our experiments. Consequently, we promote sparsity in our feature representation. In autoencoders, encoding and decoding weights are often tied, which means that the decoder weight matrix is the transpose of the encoder weight matrix: $W_D = W_E'$. This reduces the amount of free parameters available, but also the expressive power of the model. In our experiments, instead, we optimise $W_D$ and $W_E$ independently. This makes our model more expressive at the cost of more computational overhead and possible delayed convergence. Another aspect that increases the computational cost of our model is the use of sparse autoencoders as opposed to autoencoders with bottleneck architecture which have fewer nodes in hidden layer and, consequently, reduced memory and computational complexity.
However, the computational cost is linear in the number of training samples, and thus it is more efficient than graph based semi-supervised learning algorithms which have cubic complexity $O(N^{3})$.

\section{EXPERIMENTS}
\label{sec:experiments}

\subsection{Experimental Setup}
We performed our experiments on the standard TIMIT data set~\cite{timit} for frame-based phoneme classification. We used the standard core test set of 192 sentences, and a development/validation set of 184 sentences. For training, we had 3512 sentences. Similarly as a part of standard procedure of experiments on TIMIT, glottal stop segments are excluded. 
The data is created with the help of standard recipes given in~\cite{kaldi, pdnn}. The input to our network was created by first extracting a 39-dimensional feature vectors for each frame. The feature vector is made of 12 MFCC coefficients computed at a rate of 10 ms with an overlapping window of 20 ms, 1 energy coefficient, deltas and delta-deltas. For each time step, the features obtained 5 frames to the left to 5 frames to the right are concatenated together to form a final vector has a dimension of $11\times 39= 429$ coefficients as in~\cite{mohamed-2009}. Speaker-dependent mean and variance normalisation was also applied.

The total number of frames in the training set is 1068816. The validation set has 56005 frames in total, and the test set has 57919 frames. These counts are in line with the experiments of~\cite{liu-2013, labiak-2011}.
For training, we used the standard phone set of 48 phones, collapsed into 39 phones for evaluation as in~\cite{lee-1989}. This means, the output layer will have 48 nodes, but at the time of evaluation, the 48 phonemes will be reduced to 39 phonemes. 
This procedure has also been used in~\cite{liu-2013}.
Although it is more common to use senones as the target labels for the classification network as in~\cite{mohamed-2009}, the output of our classification network was based on phonemes in order to be able to compare with other studies on semi-supervised learning for speech.

To simulate the effect of missing labels during training, the training set was divided into a labelled portion and an unlabelled portion of data set. The percentage of labelled frames in the training set was varied from 1\% to 30\% with intermediate steps: 3\%, 5\%, 10\%, 20\%. For each of these conditions, we optimised the hyper-parameter $\alpha$ on the validation set. All the accuracy results are reported for the optimal value of $\alpha$. Finally the number of nodes in the encoder network was also optimised on the validation set resulting in an optimal value of 10000 nodes.

As a baseline, we compare the results obtained with our method with those obtained with a similar neural network trained with supervised backpropagation, on the same amount of labelled examples.
We also compare our results with those obtained in the literature on semi-supervised learning.

\begin{figure}
\includegraphics[width=\columnwidth]{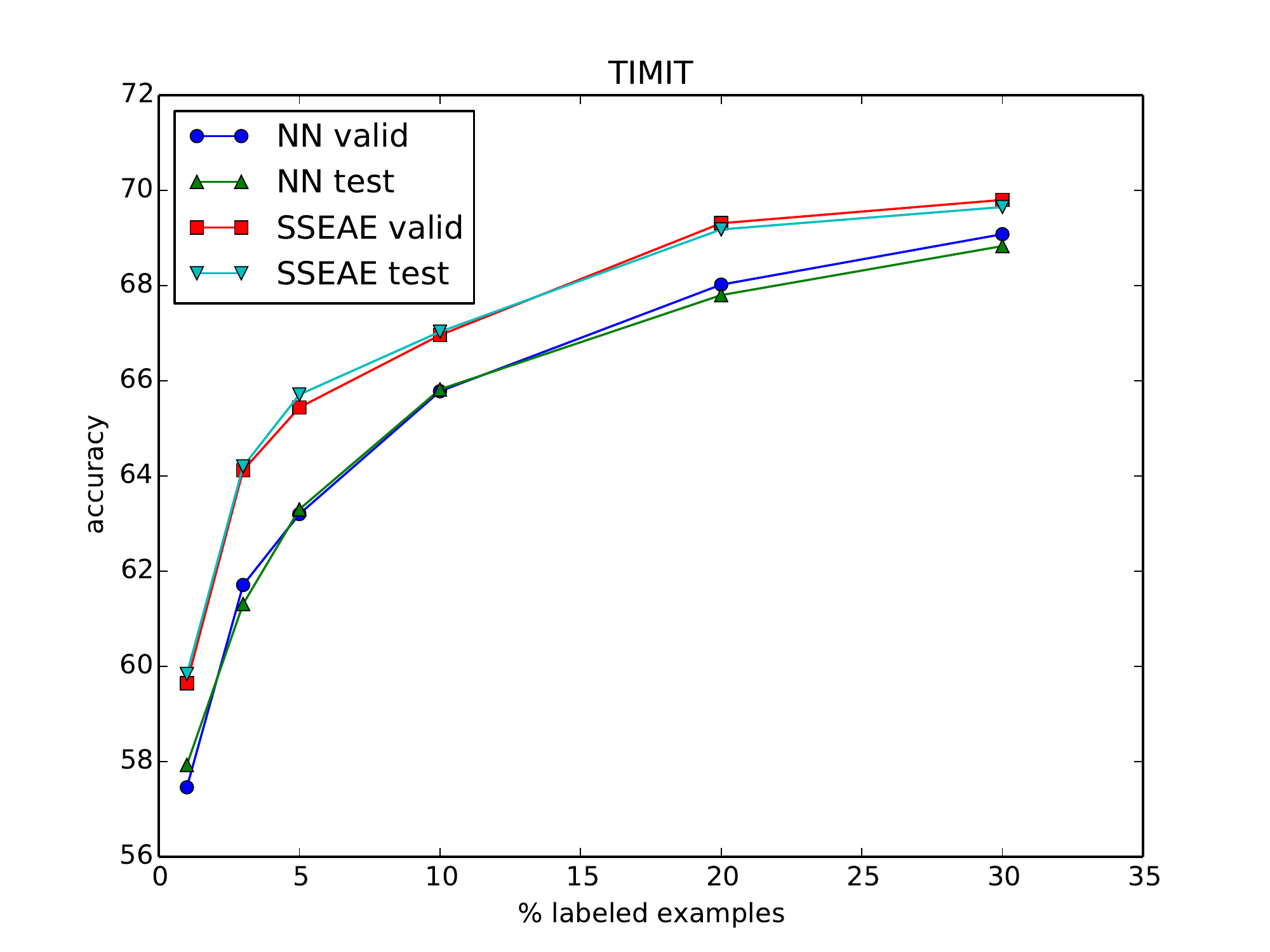}
\caption{Frame-based phoneme recognition accuracy (\%) versus percentage of labelled training examples on the TIMIT database. NN: neural network trained with supervised backpropagation. SSSAE: our method. Both validation and test accuracy rates are shown. See Table~\ref{table:timit} for the corresponding numerical values.}
\label{fig:results}
\end{figure}

\subsection{Practical Setup}
We used Kaldi~\cite{kaldi} and PDNN~\cite{pdnn} for feature extraction, Theano~\cite{theano} for symbolic algebra and GPU computing. The experiments were run on a Titan X card installed on a Ubuntu 14.04 based machine.


\begin{table*}
\centering
  \begin{tabular} {rrccccc}
    \hline\hline
    \multicolumn{7}{c}{Results on TIMIT} \\
    \hline
   \multicolumn{2}{c}{Labelled Observations} & \multicolumn{2}{c}{Neural Network} & \multicolumn{2}{c}{SSSAE} & \\
   \% & \# & valid. acc. (\%) & test acc. (\%) & valid acc. (\%) & test acc. (\%) & $\alpha$ \\
    \hline
  1 & 10688 & 57.46 & 57.93 & 59.65 & 59.84 & 100 \\
  3 & 32065 & 61.71 & 61.31 & 64.12 & 64.20 & 150 \\
  5 & 53441 & 63.20 & 63.30 & 65.44 & 65.71 & 150 \\  
  10 & 106881 & 65.78 & 65.82 & 66.96 & 67.03 & 400 \\
  20 & 213763 & 68.02 & 67.80 & 69.31 & 69.18 & 600 \\
    30 & 320644 & 69.08 & 68.83 & 69.80 & 69.65 & 900 \\
    \hline\hline
  \end{tabular}
    \caption{Results on frame-based phoneme classification on the validation and test sets on the TIMIT material. Our method (SSSAE) is compared to a neural network trained with supervised backpropagation with the same amount of labelled data. The total number of training frames is 1068818. The value of $\alpha$ is optimised on the validation set as the proportion of labelled examples is varied.}
    \label{table:timit}
\end{table*}

\section{RESULTS}
\label{sec:results}
Table~\ref{table:timit} and Figure~\ref{fig:results} show the frame-level classification accuracy rates for a neural network trained in a supervised way (NN) and the proposed single layer semi-supervised sparse auto-encoder (SSSAE) for varying percentage of labelled data. Both validation set accuracy and test set accuracy are reported. The hyper-parameters of the neural network such as learning rate tuned using the validation set are also shown in the table. The neural network contains 2000 units in the hidden layer as in~\cite{liu-2013,labiak-2011} and performs similarly to the one reported there.

Table~\ref{table:timit} and Figure~\ref{fig:results} show that the method always performs better than the supervised baseline by as much as 2.9\% absolute improvement. As expected this advantage decreases when the proportion of labelled training examples is increased.
The validation and test errors are always very close, indicating that the parameters optimised on the validation set generalise well to the test set.
As expected, the optimal value for $\alpha$ is strongly dependent on the proportion of labelled material. The higher the proportion the more weight the algorithm gives to the classification error, compared to the unsupervised reconstruction error.

In Table~\ref{table:timitssl}, we compare the performance of our system to the results obtained with graph based semi-supervised learning methods published in \cite{liu-2013} on 10\% and 30\% labelled data.
We observe that our system performs better than all the techniques mentioned except the Prior Regularised Measure Propagation (pMP) algorithm. 
\begin{table}
\centering
  \begin{tabular}{lccc}
    \hline\hline
    \multicolumn{4}{c}{Comparison with other methods} \\
    \hline
    & & 10\% labelled & 30\% labelled \\
    Method & Reference & \multicolumn{2}{c}{Test accuracy (\%)} \\
    \hline
    NN & this work & 65.94 &  69.24 \\
    LP & \cite{liu-2013} & 65.47 & 69.24 \\
    MP & \cite{liu-2013} & 65.48 & 69.24 \\
    MAD & \cite{liu-2013} & 66.53 &  70.25 \\
    pMP & \cite{liu-2013} & 67.22 & 71.06 \\
  SSSAE & this work & 67.03 & 69.65 \\
    \hline\hline
  \end{tabular}
    \caption{Accuracy rates (\%) for frame-based phoneme classification on TIMIT for the baseline (NN), the four different algorithms in GBL-SSL \cite{liu-2013} and our model, SSSAE}
    \label{table:timitssl}
\end{table}


\section{CONCLUSIONS}
\label{sec:conclusions}
We reported results on frame based phoneme classification on the TIMIT database using semi-supervised learning based on sparse autoencoders.
We observe that our method outperforms a neural network trained with supervised backpropagation on the same amount of labelled training data in all experimental conditions. 
Our results also outperform many of the semi-supervised learning methods proposed in the literature for a similar task, with the exception of Prior-Regularised Measure Propagation (pMP) method.
As expected, the advantage of using our method decreases when the proportion of labelled training observations is increased.
However, we can argue that in realistic situations we will always find an abundance of unlabelled data as compared to data that was carefully annotated.
As a consequence, it becomes more important for us to investigate our model when the percentage of labelled data is low.

In spite of the promising results, in order to draw general conclusions on ASR, we would need to test our method on a word recognition task, and, in particular, on large-vocabulary ASR. However, the improvements we see in frame-level phoneme classification are an incentive to continue work in this direction.
Possible improvements may be obtained by using alternative features (e.g. filterbank features) instead of the MFCCs that were used here to allow for comparison with previous results in the literature.
We may also test how the results vary if we add depth to the model, by stacking several blocks of autoencoders/classifiers.



\section{ACKNOWLEDGMENTS}
The GeForce GTX TITAN X used for this research were donated by the NVIDIA Corporation.

\label{sec:refs}

\bibliographystyle{IEEEbib}
\bibliography{strings,refs}

\end{document}